\documentclass[10pt,twocolumn,letterpaper]{article}

\usepackage[review]{cvpr}      

\usepackage{graphicx}
\usepackage{amsmath}
\usepackage{amssymb}
\usepackage{booktabs}

\usepackage[pagebackref,breaklinks,colorlinks]{hyperref}

\usepackage[capitalize]{cleveref}
\crefname{section}{Sec.}{Secs.}
\Crefname{section}{Section}{Sections}
\Crefname{table}{Table}{Tables}
\crefname{table}{Tab.}{Tabs.}

\usepackage{float} 
\usepackage{dirtytalk} 
\usepackage{algorithmicx}
\usepackage{algorithm}
\usepackage[noend]{algpseudocode} 

\usepackage{xpatch}

\xpatchcmd{\algorithmic}{\setcounter}{\algorithmicfont\setcounter}{}{}
\providecommand{\algorithmicfont}{}

\def\cvprPaperID{05} 
\def\confName{CVPR}
\def\confYear{2022}

\usepackage{overpic}
\usepackage{enumitem} 
\usepackage{overpic} 
\usepackage{color}

\definecolor{turquoise}{cmyk}{0.65,0,0.1,0.3}
\definecolor{purple}{rgb}{0.65,0,0.65}
\definecolor{dark_green}{rgb}{0, 0.5, 0}
\definecolor{orange}{rgb}{0.8, 0.6, 0.2}
\definecolor{red}{rgb}{0.8, 0.2, 0.2}
\definecolor{darkred}{rgb}{0.6, 0.1, 0.05}
\definecolor{blueish}{rgb}{0.0, 0.3, .6}
\definecolor{light_gray}{rgb}{0.7, 0.7, .7}
\definecolor{pink}{rgb}{1, 0, 1}
\definecolor{greyblue}{rgb}{0.25, 0.25, 1}

\usepackage{blindtext}

\renewcommand{\paragraph}[1]{\vspace{1em}\noindent\textbf{#1}.}

\usepackage{amsmath,amsfonts,bm}

\def\eqref#1{equation~\ref{#1}}

\def\1{\bm{1}}

\def\vx{{\bm{x}}}

\DeclareMathAlphabet{\mathsfit}{\encodingdefault}{\sfdefault}{m}{sl}
\SetMathAlphabet{\mathsfit}{bold}{\encodingdefault}{\sfdefault}{bx}{n}

\newcommand{\R}{\mathbb{R}}

\begin{document}

\title{Guided Deep Metric Learning}


\author{ {\small Jorge Gonzalez-Zapata$\textsuperscript{1}$, Ivan Reyes-Amezcua$\textsuperscript{1}$, Daniel Flores-Araiza$\textsuperscript{2}$, Mauricio Mendez-Ruiz$\textsuperscript{2}$,}\\ {\small Gilberto Ochoa-Ruiz$\textsuperscript{2}$, Andres Mendez-Vazquez$\textsuperscript{1}$}\\\\
{\small $\textsuperscript{1}$CINVESTAV Unidad Guadalajara, Mexico}\\
{\small $\textsuperscript{2}$Tecnológico de Monterrey, School of Engineering and Sciences, Mexico}\\
{\tt\small jorge.gonzalezzapata@cinvestav.mx, gilberto.ochoa@tec.mx, andres.mendez@cinvestav.mx}
}
\maketitle

\begin{abstract}
Deep Metric Learning (DML) methods have been proven relevant for visual similarity learning. However, they sometimes lack generalization properties because they are trained often using an inappropriate sample selection strategy or due to the difficulty of the dataset caused by a distributional shift in the data. These represent a significant drawback when attempting to learn the underlying data manifold. Therefore, there is a pressing need to develop better ways of obtaining generalization and representation of the underlying manifold. In this paper, we propose a novel approach to DML that we call \textit{Guided Deep Metric Learning}, a novel architecture oriented to learning more compact clusters, improving generalization under distributional shifts in DML. This novel architecture consists of two independent models: A multi-branch master model, inspired from a Few-Shot Learning (FSL) perspective, generates a reduced hypothesis space based on prior knowledge from labeled data, which guides or regularizes the decision boundary of a student model during training under an offline knowledge distillation scheme. Experiments have shown that the proposed method is capable of a better manifold generalization and representation to up to 40\% improvement (Recall@1, CIFAR10), using guidelines suggested by Musgrave et al. to perform a more fair and realistic comparison, which is currently absent in the literature.
\end{abstract}
\section{Introduction}
\label{sec:intro}
\noindent
DML has proven to be a relevant topic given its strategy of acting directly on the resulting embedding distances to capture the semantic similarity of the data, using the robustness of deep learning models. Over time, DML methods have been integrated to task such as zero-shot \cite{roth2021simultaneous, milbich2021characterizing}, few-shot \cite{tian2019contrastive, tian2020rethinking} and self-supervised learning \cite{rajasegaran2020self,ji2021power, milbich2020diva}.

Baseline DML methods have a variety of proposals with different loss functions \cite{kaya2019deep,deng2019arcface}. However, pair or triplet samples imply high complexity time in the training process. Thus, were introduced sample mining strategies \cite{kaya2019deep, harwood2017smart}. Still, these strategies do not generalize well to all architectures and can be counterproductive depending on the nature of the data or architecture \cite{wu2017sampling,musgrave2020metric}.
In addition, the experiments realized in \cite{roth2020revisiting} and \cite{musgrave2020metric} indicate overall flaws in the experimental setups in DML that lead to unfair comparison. 

Most Machine Learning (ML) models usually assume that the train and test data are drawn from the same distribution (i.i.d. assumption). However, in a realistic scenario, distributional shifts between the train and test set can occur, where the test distribution is unknown and diverges from the train distribution. Precisely, Out-Of-Distribution (OOD) generalization address this problem \cite{shen2021towards}.

While there is still a vague definition of OOD in the literature and characterization of distributional shift is still an open problem \cite{shen2021towards}, we have opted to use these concepts using several references \cite{yang2021generalized,koh2021wilds,milbich2021characterizing,shen2021towards} that remain congruent in certain properties. For example, the distributional shifts can be caused by \textit{semantic shift} (or label shift), in which the OOD samples belong to new classes, or by \textit{covariate shift}, where the distribution of the data changes between the train and test scenarios while keeping the same labels. Our proposal focuses on solving for covariate shift.

Among the causes of covariate shift are included problems related to \textit{domain generalization}, i.e., when the train and test domains are disjoint but still share the same labels. As well as by problems related to \textit{sub-population shift}, i.e., when the train and test domains are the same, but their proportions are different. Ideally, a DML model learns an embedding space that generalizes well enough within the train data distribution, known as In-Distribution (ID), avoiding vulnerabilities to data difficulty (unspecific covariate shift). Some methods make use of diverse concepts such as \textit{knowledge distillation} \cite{furlanello2018born,kim2021embedding,roth2021simultaneous} and consider a probable distributional shift in the data \cite{milbich2021characterizing,du2022vos} to improve generalization. 

For example, \cite{milbich2021characterizing} show through exhaustive experiments how data splitting into train and test involves different distribution shifts that modify the difficulty of the data and proposes few-shot DML to improve generalization upon unknown shifts in test.

\begin{figure}[ht!]
\centering
\captionsetup{justification=justified}
\includegraphics[width=0.5\textwidth]{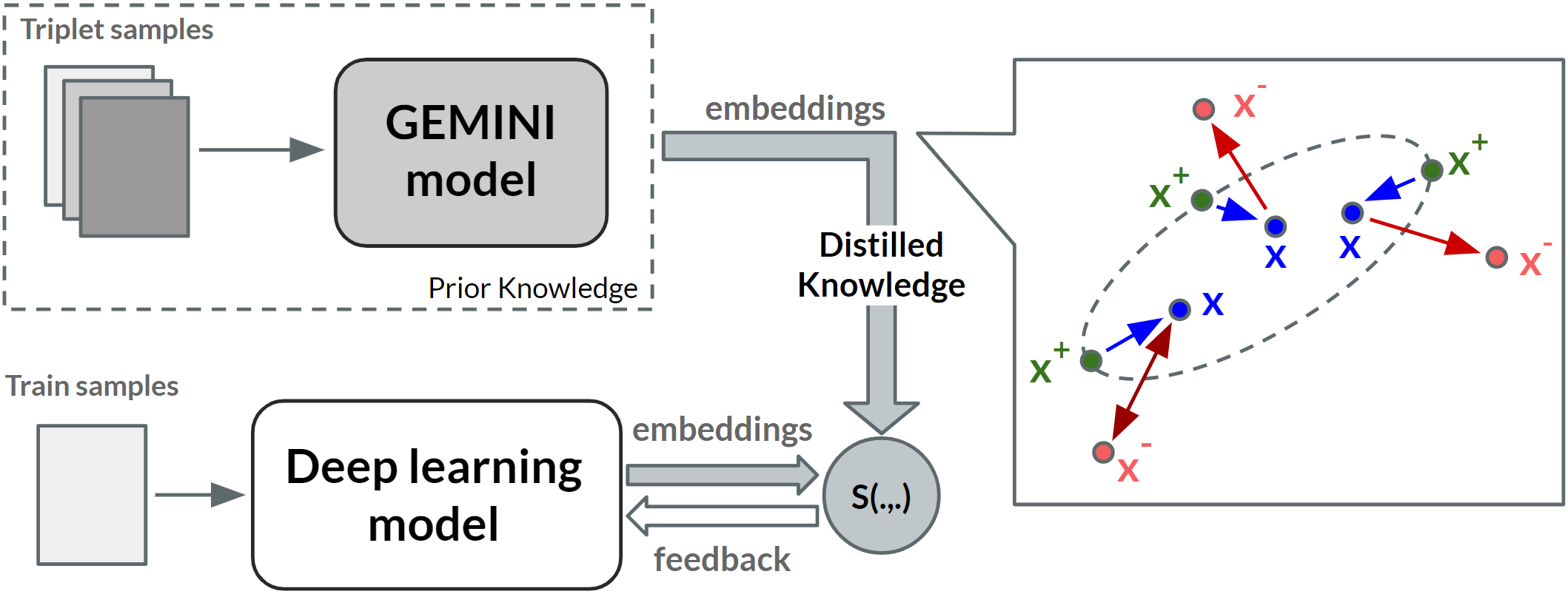}
\caption{Block diagram of our proposal. The reduced hypothesis space generated a priori by the GEMINI model is used to guide the training of the deep learning model.}
\label{fig:propBlocks}
\end{figure}

This paper presents an approach that brings together adaptations of both Few-Shot Learning (FSL) and knowledge distillation that avoids the restrictions of classification layers. The proposal consists of two independent models: The first one is a multi-branch master model, called GEMINI, that exploits local and global information to generate a reduced hypothesis space based on prior knowledge from the source labeled dataset in the form of triplet samples. This model represents a fast-convergence compact model, with negligible train time cost and no change in test time cost, avoiding problems of large teacher networks and time consumption \cite{furlanello2018born}. Its architecture is an analogy to strategies based on parameter sharing in FSL \cite{wang2020generalizing}. The second model, a deep learning model, is a student model that learns an adequate embedding function guided by the mentioned reduced hypothesis space by using a similarity function $s(\cdot,\cdot)$, following the teacher-student concept from knowledge distillation \cite{hinton2015distilling,furlanello2018born}. Exploiting the tractability of the features space low-dimensionality to regularize the student decision bounds (Figure \ref{fig:propBlocks}).

To test our proposal, we have followed some of the guidelines proposed by \cite{musgrave2020metric} to design an experimental protocol that allows us to compare different models under equal conditions as much as possible. Our proposal demostrate the following key contributions:
\begin{itemize}
  \item The experiments empirically show that the quality of the GEMINI embedding space is reliable given the consistency of the results with a random sample selection, reducing the dependence on the not-so-reliable sample mining strategies.
  \item Following \cite{milbich2021characterizing}, we further show that FSL adaptations in DML improve generalization with learning circumstances hindered by distributional shifts.
  \item The evaluation metrics suggest that our proposal can obtain better-delimited embedding spaces with compact clusters than with the compared models, giving less uncertainty between ID and OOD data.
  \item The performance in low-dimensional (two-dimensional) embeddings positions the proposed model as useful for data visualization.
\end{itemize}

\section{Related works}
\label{sec:related}
\noindent
\textbf{Deep Metric Learning.} Usually, there are two fundamental DML models considered in the literature: The former is the \textit{siamese network} (contrastive loss) \cite{chopra2005learning,hadsell2006dimensionality}, a method based on pairwise samples, which encourages small positive pairwise distances and negative pairwise distances above a certain margin. The latter is the \textit{triplet network} (triplet loss) \cite{hoffer2015deep} that considers three types of samples: positive, negative, and an anchor. In this case, the distance between the anchor-negative samples should be greater than the distance between the anchor-positive samples by at least a margin. The triplet network improves over the siamese network by using intra-class and inter-class relationships \cite{musgrave2020metric}, allowing a better fit to the variance differences between classes and making the model less restrictive. In this way, using these same fundamental architectures, there are a variety of proposed loss functions, e.g., Angular loss \cite{wang2017deep}, Mixed loss \cite{chen2018dress}, Margin Loss \cite{wu2017sampling}, Multi-similarity loss \cite{wang2019multi}, N-Pairs loss \cite{sohn2016improved}, among others \cite{kaya2019deep,deng2019arcface}. 

However, pairwise or triplet samples can involve high time complexity in the training process. Hence, \textit{sample mining} strategies to identify the most informative examples capable of increasing performance, as well as the training speed, can be used. Nevertheless, for instance, in the case of hard-negative mining, the siamese network generally converges faster. However, the case of the triplet network often leads to problems where all samples have the same embedding and produce noisy gradients \cite{wu2017sampling,musgrave2020metric}. Meanwhile, semi-hard negative mining is recommended for triplet network over hard-negative mining to avoid the risk of \textit{overfitting}. However, in some cases, it might converge quickly at first, but as the number of negative examples within the margin runs out, it drastically slows down its progress \cite{wu2017sampling}. These indicate that choosing an appropriate sampling strategy could be a difficult decision, as it seems to be sensitive to the properties of the underlying dataset or when architecture changes occur.

Meanwhile, new approaches have extended DML methods to other topics and have shown improvements in leveraging data relationships. Such is the case of Zero-Shot Learning (ZSL) and Few-Shot Learning (FSL), paradigms that address applications hindered by a limited number of samples and where it uses prior knowledge to generalize quickly \cite{wang2020generalizing,milbich2021characterizing}.

The ZSL approach, where test and train classes are distinct, intends to learn representation spaces that capture and transfer visual similarity to unseen classes \cite{milbich2021characterizing,roth2021simultaneous,brattoli2020rethinking,oh2016deep}. Faces the challenge of constructing a priori unknown test distribution with an unspecified distributional shift from the train distribution. However, arbitrarily large distributional shifts may cause the captured knowledge from the training data to be less significant to the test data \cite{milbich2021characterizing}, i.e., ill-posed learned representations.
In the case of FSL, where at least a few samples of the test distribution are available during training, improve the quality of embeddings or prototype representations \cite{tian2020rethinking,rajasegaran2020self,sung2018learning,snell2017prototypical}. Specifically, in \cite{milbich2021characterizing} has been proven that adaptations of FSL can improve the generalization capability of DML since even the minor additional domain knowledge provided helps to adjust the learned representation space to achieve better OOD generalization (commonly referring to covariate shift).

Also related to our line of research, there are some approaches following FSL methods that learn by constraining the hypothesis space through prior knowledge using a \textit{parameter sharing} strategy \cite{wang2020generalizing}. Such is the case of fine-grained image classification \cite{zhang2018fine}, domain adaptation \cite{motiian2017few}, and cross-domain translation \cite{benaim2018one}, where some layers are for capturing global information and others for local information.\\

\noindent
\textbf{Knowledge Distillation.} Originally introduced as model compression and acceleration techniques, \textit{knowledge distillation} refers to an approach with teacher-student architecture \cite{hinton2015distilling,gou2021knowledge}. The main idea is to learn a student model (small network) from a teacher model (large network). As a result, it's obtained a small network trained to learn to replicate the behavior of the original network. This idea evolved beyond the goal of model compression and was subsequently used to improve performance in computer vision and language modeling tasks \cite{furlanello2018born}. Specifically, in the context of DML, the purpose is to take advantage of the knowledge captured by the master model to learn better embedding functions \cite{chen2018darkrank,tian2019contrastive,tian2020rethinking}.

In addition, there are variants within this approach with more specific configurations, such as \textit{Self-distillation} \cite{zhang2019your,roth2021simultaneous}, where the same network acts as both teacher and student models. As well as \textit{self-supervised learning} \cite{rajasegaran2020self,milbich2020diva}, oriented to help initialize a network where there is a lack of labeled data. In self-supervised, an initial (pretext) task is learned by a master model using a general content dataset (e.g., ImageNet), which provides the \say{supervision} to the student model to perform the actual (downstream) task. A common approach in self-supervised learning uses \textit{contrastive learning} to act over the similarity between the embeddings of pair samples \cite{ji2021power}.

\begin{figure*}
\begin{center}

\includegraphics[width=0.7\textwidth]{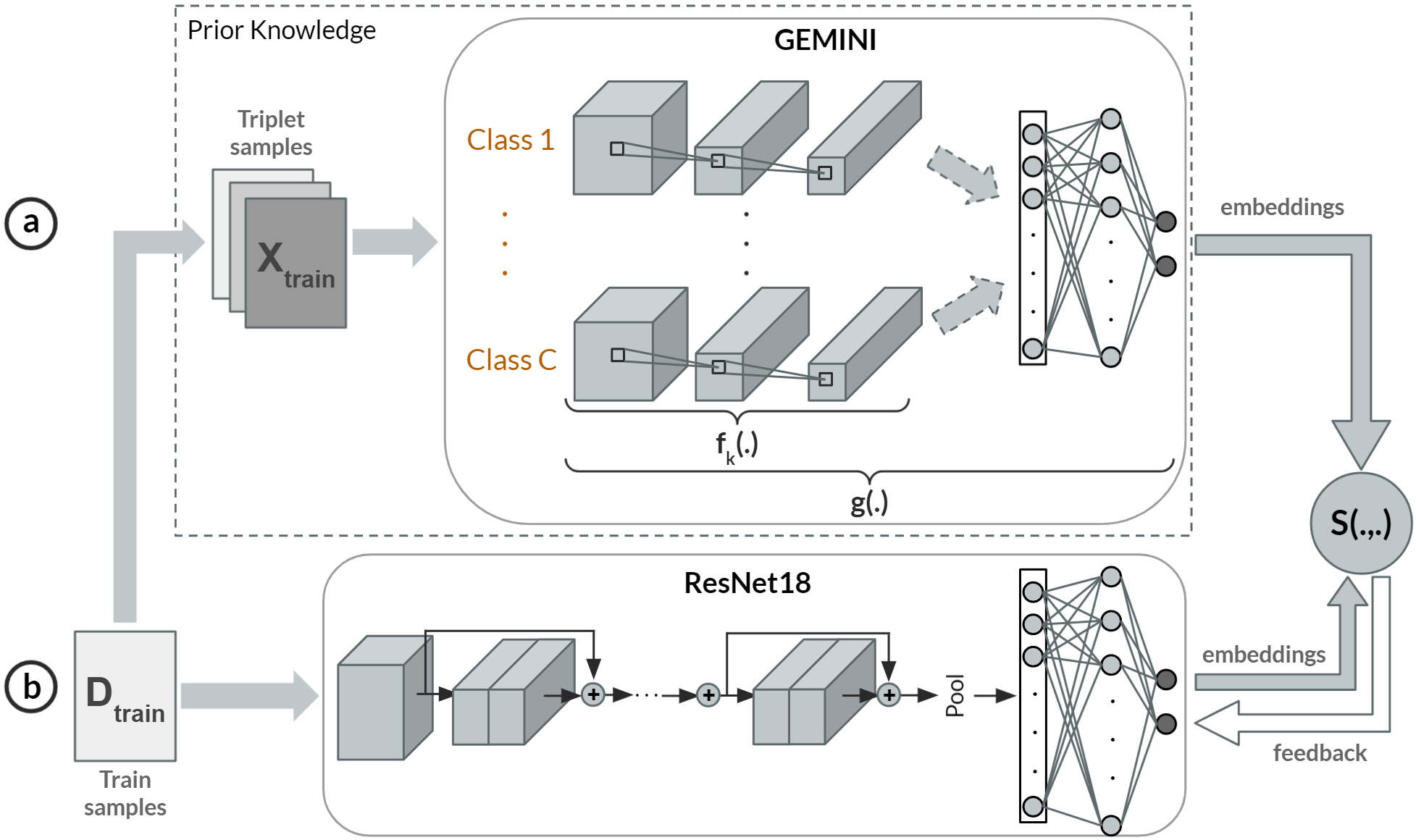}

\end{center}
\caption{Complete Architecture. First, \textbf{a)} The GEMINI model is trained using the triplet data samples $X_{train}$ generated from the original train dataset $D_{train}$. Then, we acquire the embeddings of the respective train dataset $D_{train}$ from the reduced hypothesis space generated. Second, \textbf{b)}  The GEMINI's low-dimensional embeddings guide the training of the ResNet through a similarity measure $S(\cdot,\cdot)$ in this low-dimensional space, comparing the distance between the respective embeddings. The discrepancy provides feedback for the ResNet parameters.}
\label{fig:complete}
\end{figure*} 

\section{Method}
\noindent
In this section, we propose a novel architecture for the task of deep metric learning, called \textit{Guided deep metric learning}. This architecture has key concepts based on both FSL and Knowledge distillation. This architecture consists of two independent models (embedding functions). The first is called \textit{GEMINI}, a fast-convergence compact model, which is in charge of generating a reduced hypothesis space based on prior knowledge. The second one is a deep learning architecture (specifically, a ResNet-18) used to learn the embedding function guided by the hypothesis space generated by the GEMINI model.

\subsection{GEMINI Model}
\noindent
Analogous to parameter sharing strategies found on FSL, the GEMINI model consists of two main parts (Figure \ref{fig:complete}). The first component $f_k(\cdot)$ exploits the local information of each class, one stream layer \cite{aghamaleki2019multi,chenarlogh2019multi} per class. Then, the global fully connected component $g(\cdot)$ tries to exploit the global information of local representations by sharing some parameters between the different classes, producing a unique data representation and, thus, avoiding the strong restrictions of a classification layer, e.g., cross-entropy.

The model uses a triplet dataset $X$ generated from a training dataset $D_{train}$ with $c$ classes. We will use the notation $\vx_{(k)}$ and $\vx^{+}_{(k)}$ for the anchor and positive samples of class $k$ and $\vx^{-}_{(l)}$ for the negative sample of class $l$, where $k \neq l$. Each sample is input to the network through its respective stream, depending on its class. The outputs are intermediate representations denoted by $f_i(\vx^{*}_{(i)})$ for each class $i = 1 \dots c$, regardless of the sample type ($^*$).

\begin{equation}
    f_{k}\left(\vx_{\left(k\right)}^{*}\right) = h_{\left(k\right)}^{L}\circ h_{\left(k\right)}^{L-1}\circ\cdots\circ h_{\left(k\right)}^{1}\left(\vx_{\left(k\right)}^{*}\right)
    \label{LayerStream}
\end{equation}

\noindent
The equation (Eq. \ref{LayerStream}) represents the layer composition performed at each stream. Here, the triplets $(\vx_{(k)},\vx^{+}_{(k)},\vx^{-}_{(l)})$ are selected in two steps: First, a permutation of length 2 (without replacement) is randomly selected from the classes, thus selecting two respective streams. Second, two samples of the first class ($\vx_{(k)},\vx^{+}_{(k)}$) and one of the second class ($\vx^{-}_{(l)}$) are extracted from the selected permutation. 

Each mini-batch contain a certain number of triplet samples, given by a hyper-parameter, of one permutation of the classes, this is denoted by $X_{b}$ where $b = 1, \dots, \frac{\left|c\right|!}{(\left|c\right|-2)!}$. This means that, given that there are only two different classes in each mini-batch, only two stream are used (activated) in each mini-batch. Thus, the activation of the streams are control by the given indicator function $\textbf{1}_{ij}$, where $i$ and $j$ are the stream index and the sample class respectively, 
\begin{equation} \label{indfunc}
\textbf{1}_{ij} = \left\{
\begin{array}{ll}
      1, & \textrm{if }  i = j \\
      0, & \textrm{Otherwise}\\
\end{array} 
\right.
\end{equation}

\noindent
In order that, during the training process, the following condition is enforced,
\begin{equation} \label{streamsfunc}
    \textbf{1}_{ij} \cdot || f_{i}( \vx_{(j)}) - f_{i}( \vx^{+}_{(j)} ) || < \epsilon \textrm{,} \quad \forall \left(\vx_{(j)},\vx^{+}_{(j)}\right) \in X_{b}
\end{equation}
\begin{equation} \label{streamsfunc2}
    \textbf{1}_{il} \cdot f_{i}(\vx^{-}_{(l)}) \textrm{,} \quad \forall \vx^{-}_{(l)} \in X_{b}
\end{equation}

\noindent
where $\epsilon$ is a small positive number. This is the expected behaviour in each update of the model parameters after each mini-batch, in accordance with the proposed cost function (Eq. \ref{costfunc}). However, once they passed through the local component $f_k(\cdot)$, they all pass through the global component $g(\cdot)$ where the samples in the mini-batch share the layer parameters. The embedded representation of the network is denoted by $g(\vx^{*}_{(i)}) = g(f_i(\vx^{*}_{(i)}))$. Thus, once the intermediate and final representations are obtained for each sample in the triplet sample, both components of the network are coupled through the proposed cost function (Eq. \ref{costfunc}).
\begin{align}
L\left(f, g, d, \beta, M\right) & =  \frac{1}{\left| X_{b} \right|} \sum_{\left(\vx_{(k)},\vx^{+}_{(k)},\vx^{-}_{(l)}\right) \in X_{b}} \nonumber \\
& \beta \cdot d\left( f_k \left(\vx_{(k)}\right), f_k \left(\vx^{+}_{(k)}\right)\right) + ... \nonumber \\ 
&  (1-\beta) \left[ M - d\left(g\left(\vx_{(k)}\right), g\left(\vx^{-}_{(l)}\right)\right)\right]_{+} 
\label{costfunc}
\end{align}
\noindent
where the term $\beta \in \left[0,1\right]$ is a weighting parameter, the term $[x]_{+}=max(0,x)$ is the hinge loss, and $M$ is a margin. The first term of the cost function evaluates the closeness of the similar samples, using the local information of the class to emphasize the closeness of the intermediate representations. 

Note that using this first term alone may certainly lead to a trivial solution, i.e., a mapping of all points to a single point in the embedded space. To prevent this type of solution, a second term has been added which sets away points belonging to different classes. Thus, as the training progress, only the pairs that satisfy $ d(g(\vx_{(k)}), g(\vx^{-}_{(l)})) < M $ will produce a cost value. In this way, the first term minimizes intra-class distances, and the second term prevents trivial solutions by maximizing the distances between classes, inter-class distances.

The term $M$ can be defined as $M = d(g(\vx_{(k)}), g(\vx^{+}_{(k)})) + m$, where $m$ is a margin value. This makes the second term of the cost function resembles the triplet loss function, but with an added term that keeps the similar samples together using the local information. To balance both effects in the loss function there is a $\beta$ term which, experimentally, its value is usually small ($\approx 0.005$), but significant to  improve the resulting reduced hypothesis space.\\

\begin{algorithm}[h]
\caption{GEMINI Model}\label{alg:cap}
\hspace*{\algorithmicindent} \textbf{Input:} original training set $D_{train} = \{(x_i,y_i)\}^{N_{train}}_{i=1}$\\
\hspace*{\algorithmicindent} \qquad \quad batch size $N_{batch}$\\
\hspace*{\algorithmicindent} \qquad \quad number of classes $C$\\
\hspace*{\algorithmicindent} \qquad \quad output dimension size $OutputSize$\\
\hspace*{\algorithmicindent} \qquad \quad number of epochs $N_{epochs}$\\
\hspace*{\algorithmicindent} \qquad \quad weighting parameter $\beta$\\
\hspace*{\algorithmicindent} \qquad \quad margin value parameter $m$\\
\hspace*{\algorithmicindent} \textbf{Output:} reduced hypothesis space embeddings $\hat{Z}$
\begin{algorithmic}[1]
\State $X_{train} \gets MakeTripletSamples\left(D_{train}, N_{batch}\right)$
\State $f \gets [f_1, \cdots, f_{C}]$
\State $g \gets FullyConnected(OutputSize)$

\For{$i \gets 1 \textrm{ to } N_{epochs}$}
\For{each  $\{(x_{(k)}, x^{+}_{(k)}, x^{-}_{(l)}),(k,l)\}^{N_{batch}}_{i=1} \in X_{train}$}
       \State $a \gets f_k(x_{(k)})$
       \State $p \gets f_k(x^{+}_{(k)})$
       \State $anchor \gets g(a)$
       \State $positive \gets g(p)$
       \State $negative \gets g(f_l(x^{-}_{(l)}))$
       \State $loss \gets \beta \cdot||a-p|| + (1-\beta)(||anchor-positive||+m-||anchor-negative||)$
       \State $\textrm{GEMINI}.gradient.step(loss)$
\EndFor
\EndFor
\State $\hat{Z} \gets \textrm{GEMINI}(D_{train})$
\end{algorithmic}
\end{algorithm}

\noindent
The model works by decreasing the distances between anchor and positive samples and, simultaneously, increasing the distance between anchor and  negative samples (Figure \ref{fig:propBlocks}). Rather than engaging with sample mining techniques, which, as mentioned before, have been shown not to work well for all possible scenarios, restricting the generality. An alternative approach arises when the cost function (Eq. \ref{costfunc}) is more generalized with a small change in the second term (Eq. \ref{modcostfunc}).
\begin{equation} \label{modcostfunc}
    \left[ M - d\left(g\left(\vx_{(k)}\right), \vx^{-}_{(l)} \right)\right]_{+}
\end{equation}
\noindent
In this rewriting of the second term, the transformation $g(\vx^{-}_{(l)})$ is changed by $\vx^{-}_{(l)}$. This implies that the input $\vx^{-}_{(l)}$, in this generalization, can be interpreted as a simple vector that imposes a limitation or restriction on the system. These constraints can be samples from another class, proposed by the user or even coming from another model. Thus, it can be considered that the model can shape the distribution of each class according to these specific points.

\subsection{Complete Architecture}
\noindent
\noindent
The complete architecture follows the offline knowledge distillation guidelines. In our proposal, the student network is a deep learning model of choice; in this case, we opted for the PyTorch ResNet-18 implementation. We say that this model is \say{guided} to replicate the behavior of the master network (GEMINI model). 

This approach has the advantage that GEMINI has already searched the space and arrived at a hypothesis space constrained by prior knowledge. With such a reduced hypothesis space, the ResNet model is expected to need fewer samples to converge to a suitable hypothesis, closer to the optimum, and have a lower risk of overfitting. Both models are coupled by a similarity function, $s(\cdot,\cdot)$, which measures the deviation of the ResNet hypothesis from the reduced hypothesis space (Figure \ref{fig:complete}).

The complete architecture is described in two steps: First, acquire the low-dimensionality embeddings of the reduced hypothesis space $\hat{z}_i \in Z \subseteq \R^{m}$ corresponding to each sample in the original training dataset $x_i \in D_{train} \subseteq \R^{d}$ where $m \ll d$, from a previously trained GEMINI model. Second, the ResNet model takes as input the original training dataset $D_{train}$, directly embedding each sample $x_i \in D_{train}$ into a lower-dimensional space $z_i \in Z$ without a classification layer. The resulting embeddings $z_i$ are compared with the ones obtained from the GEMINI model $\hat{z}_i$ through a similarity measure $s(z_i,\hat{z}_i)$ in the low-dimensionality space $Z \subseteq \R^{m}$. Our chosen similarity measure is the $\l_{2}$-distance. The discrepancy is then used as feedback for updating the ResNet parameters.


\begin{algorithm}[h]
\caption{Knowledge Distillation General Model}\label{alg:cap}
\hspace*{\algorithmicindent} \textbf{Input:} original training set $D_{train} = \{(x_i,y_i)\}^{N_{train}}_{i=1}$\\
\hspace*{\algorithmicindent} \qquad \quad GEMINI's reduced hypothesis space $\hat{Z}$\\
\hspace*{\algorithmicindent} \qquad \quad similarity function S($\cdot$,$\cdot$)\\
\hspace*{\algorithmicindent} \qquad \quad number of epochs $N_{epochs}$\\
\hspace*{\algorithmicindent} \qquad \quad batch size $N_{batch}$\\
\hspace*{\algorithmicindent} \textbf{Output:} low-dimensionality embeddings $Z$

\begin{algorithmic}[1]

\State $X_{train} \gets \left(D_{train}, \hat{Z} \right)$
\State $Z \gets \emptyset$

\For{$i \gets 1 \textrm{ to } N_{epochs}$}
\For{each  $\{(x, \hat{z})\}^{N_{batch}}_{i=1} \in X_{train}$}
       \State $z \gets \textrm{ResNet}.forward(x)$
       \State $loss \gets s(z,\hat{z})$
       \State $\textrm{ResNet}.gradient.step(loss)$
\EndFor
\EndFor
\State $Z \gets ResNet(X_{train})$
\end{algorithmic}
\end{algorithm}


\section{Results}
\noindent
This section evaluates, analyzes, and compares the performance of our proposed method with other approaches. The evaluation of the performance follows a few guidelines suggested by \cite{musgrave2020metric} to perform a more fair and realistic comparison. To claim that an algorithm outperforms other methods, the conditions to which the models compare must remain as similar as possible. The rationale for this approach is to ensure that it is the algorithm (not any external design choices) the one improving performance. Therefore, the designed experimental setup keeps consistency in parameter choices and avoids some commonly used techniques that could interfere with the results. For example, the use of different pre-trained network architectures (leading to differences in initial accuracies), the choice of the data augmentation strategy, and choice of the optimizer (e.g., SGD, Adam, RMSprop), among other design choices that remain inconsistent (variable) throughout the literature.

\subsection{Experimental setup}
\noindent
All of the experiments have been implemented using Python 3.8 and Pytorch 1.6 on an NVIDIA RTX 3060 super 12 GB. In addition, no data augmentation nor pre-processing (besides global normalization to zero mean and unit variance) were applied. For all the networks, the dimensionality of the output embedding space is two, gradient clipping with a factor of 0.1 and a weight decay with a decay factor of 0.0001. Further, the backbone of all networks (including our proposed method) is a ResNet-18, where the batches were constructed randomly (assuming a uniform distribution) without any sample miner. We used the repository by \cite{musgrave2020pytorch} for some losses, reducers, and DML metrics.

The networks to be compared have been trained with a learning range of 0.1-0.5, using an SGD optimizer and a batch size range of 32-512 samples. In the specific case of our proposed model, the GEMINI model has been trained at a learning rate of 0.001, margin value of 3.0, SGD optimizer with a batch size range of 32-64 triplet samples, and takes between 10 to 15 epochs to converge to a satisfactory solution. The complementary deep learning model (ResNet-18) has been trained at a learning rate of 0.1, SGD optimizer, and a batch size range of 32-128 samples.\footnote{Our source code is available at https://github.com/G-DML.}

For base performance comparison purposes, we considered the baseline models: Siamese and Triplet network, described in section 2. Additionally, recent models such as Multi-Similarity Loss \cite{wang2019multi} and Margin Loss \cite{wu2017sampling} were considered. The metrics to evaluate the performance of all architectures were the following: Recall@K, F1-score, and Normalized Mutual Information (NMI). In addition, we have used the metrics proposed by \cite{musgrave2020metric}: R-Precision (RP) and Mean Average Precision at R (MAP@R). All these metrics were obtained using the kNN classifier ($k=1$) of scikit-learn in the test embedding space (specifically, ResNet-18 output, the student model in our proposal).

The experiments were on well-known datasets: MNIST \cite{lecun1998gradient}, Fashion-MNIST \cite{xiao2017fashion}, and CIFAR10 \cite{krizhevsky2009learning}. We chose these because they are easily comparable benchmark datasets. Tables \ref{tab:comparison-stateoftheart11}-\ref{tab:comparison-stateoftheart31} show the mean performance across training runs with a 95\% confidence interval. The bold type represents the best result.

\begin{table}[H]
\centering
\resizebox{\linewidth}{!}{ 
\begin{tabular}{r|ccccc}
\hline
\textbf{Model}  & \textbf{Recall@1}             & \textbf{F-score}          & \textbf{NMI}                  & \textbf{RP}                   & \textbf{MAP@1}  \\ \hline
Siamese         & 98.27 $\pm$ 0.10              & 98.27 $\pm$ 0.10          & \textbf{95.16 $\pm$ 1.83}     & 97.88 $\pm$ 0.16              & 97.73 $\pm$ 0.17  \\
Triplet         & 97.71 $\pm$ 0.10              & 97.71 $\pm$ 0.10          & 90.98 $\pm$ 1.40              & 96.6 $\pm$ 0.10               & 96.02 $\pm$ 0.12  \\
Margin          & 98.81 $\pm$ 0.25              & 98.81 $\pm$ 0.25          & 91.56 $\pm$ 1.90              & 95.08 $\pm$ 2.88              & 96.17 $\pm$ 1.87  \\
MultiSim        & 98.58 $\pm$ 0.19              & 98.58 $\pm$ 0.19          & 90.54 $\pm$ 1.96              & 91.28 $\pm$ 1.19              & 90.12 $\pm$ 1.35  \\ \hline
Ours            & \textbf{99.00 $\pm$ 0.18}     & \textbf{99.00 $\pm$ 0.18} & 92.08 $\pm$ 1.80              & \textbf{98.56 $\pm$ 0.80}     & \textbf{98.38 $\pm$ 1.00} \\ \hline
\end{tabular}
} 
\caption{Performance on MNIST}
\label{tab:comparison-stateoftheart11}
\end{table}

\begin{table}[H]
\centering
\resizebox{\linewidth}{!}{ 
\begin{tabular}{r|ccccc}
\hline
\textbf{Model}  & \textbf{Recall@1}             & \textbf{F-score}          & \textbf{NMI}                  & \textbf{RP}                   & \textbf{MAP@1}    \\ \hline
Siamese         & 83.96 $\pm$ 0.36              & 82.89 $\pm$ 2.27          & 76.22 $\pm$ 1.40              & 80.77 $\pm$ 0.53              & 75.35 $\pm$ 0.7  \\
Triplet         & 83.86 $\pm$ 0.20              & 83.90 $\pm$ 0.21          & 75.64 $\pm$ 1.05              & 78.69 $\pm$ 0.23              & 72.07 $\pm$ 0.28  \\
Margin          & 87.75 $\pm$ 0.77              & 87.76 $\pm$ 0.77          & 79.28 $\pm$ 1.26              & 83.40 $\pm$ 2.66              & 79.91 $\pm$ 3.59  \\
MultiSim        & 90.33 $\pm$ 0.29              & 90.35 $\pm$ 0.28          & 77.30 $\pm$ 1.24              & 80.58 $\pm$ 1.56              & 77.27 $\pm$ 1.93  \\ \hline
Ours            & \textbf{91.93 $\pm$ 0.18}     & \textbf{91.91 $\pm$ 0.17} & \textbf{79.42 $\pm$ 1.66}     & \textbf{89.06 $\pm$ 0.64}     & \textbf{87.93 $\pm$ 0.66} \\ \hline
\end{tabular}
} 
\caption{Performance on Fashion-MNIST}
\label{tab:comparison-stateoftheart21}
\end{table}
\begin{table}[H]
\centering
\resizebox{\linewidth}{!}{ 
\begin{tabular}{r|ccccc}
\hline
\textbf{Model}  & \textbf{Recall@1}             & \textbf{F-score}          & \textbf{NMI}                  & \textbf{RP}                   & \textbf{MAP@1}     \\ \hline
Siamese         & 25.32 $\pm$ 1.53              & 25.96 $\pm$ 1.02          & 25.17 $\pm$ 0.99              & 24.22 $\pm$ 1.07              & 11.76 $\pm$ 4.73  \\
Triplet         & 41.93 $\pm$ 0.93              & 42.12 $\pm$ 0.95          & 39.43 $\pm$ 0.71              & 36.65 $\pm$ 0.88              & 20.03 $\pm$ 1.02   \\
Margin          & 55.71 $\pm$ 5.81              & 55.86 $\pm$ 5.84          & 45.15 $\pm$ 2.65              & 40.90 $\pm$ 3.00              & 24.73 $\pm$ 3.40   \\
MultiSim        & 56.65 $\pm$ 6.15              & 57.04 $\pm$ 6.22          & 47.51 $\pm$ 4.98              & 47.72 $\pm$ 5.96              & 36.04 $\pm$ 7.31   \\ \hline
Ours            &\textbf{80.11 $\pm$ 0.72}      &\textbf{80.09 $\pm$ 0.72}  & \textbf{61.36 $\pm$ 2.31}     & \textbf{75.98 $\pm$ 0.98}     & \textbf{73.93 $\pm$ 1.03} \\ \hline
\end{tabular}
} 
\caption{Performance on CIFAR10}
\label{tab:comparison-stateoftheart31}
\end{table}







\noindent
Given an average performance of 10 experiments for the proposed method and the other methods. As shown, in general, our proposal is superior to the others. MNIST (Table \ref{tab:comparison-stateoftheart11}) shows on average a marginal improvement. Compared to Margin loss our proposal improves by 0.19\% in Recall@1, and up to 1.3\% with respect to the Triplet network. In Fashion-MNIST (Table \ref{tab:comparison-stateoftheart21}), compared to Multi-similarity loss our proposal improves by 1.7\% and up to 9.6\% in Recall@1 compared to the Triplet network. Finally, in CIFAR10 (Table \ref{tab:comparison-stateoftheart31}) there is a clear significant improvement. Compared to Multi-similarity loss, there is an improvement of about 40\% in Recall@1 and about 3 times better compared to the Siamese network. 

In addition to better performance, our proposal consistently demonstrates stable performances (narrow confidence intervals) during the different datasets. Meanwhile, the other models showed more unstable performances, especially in the most difficult dataset (Table \ref{tab:comparison-stateoftheart31}). Thus, our proposal can be considered, in general, less sensitive to datasets with certain difficulty (distributional shift).

\begin{figure}[H]
\centering
\captionsetup{justification=justified}
\includegraphics[width=0.496\textwidth]{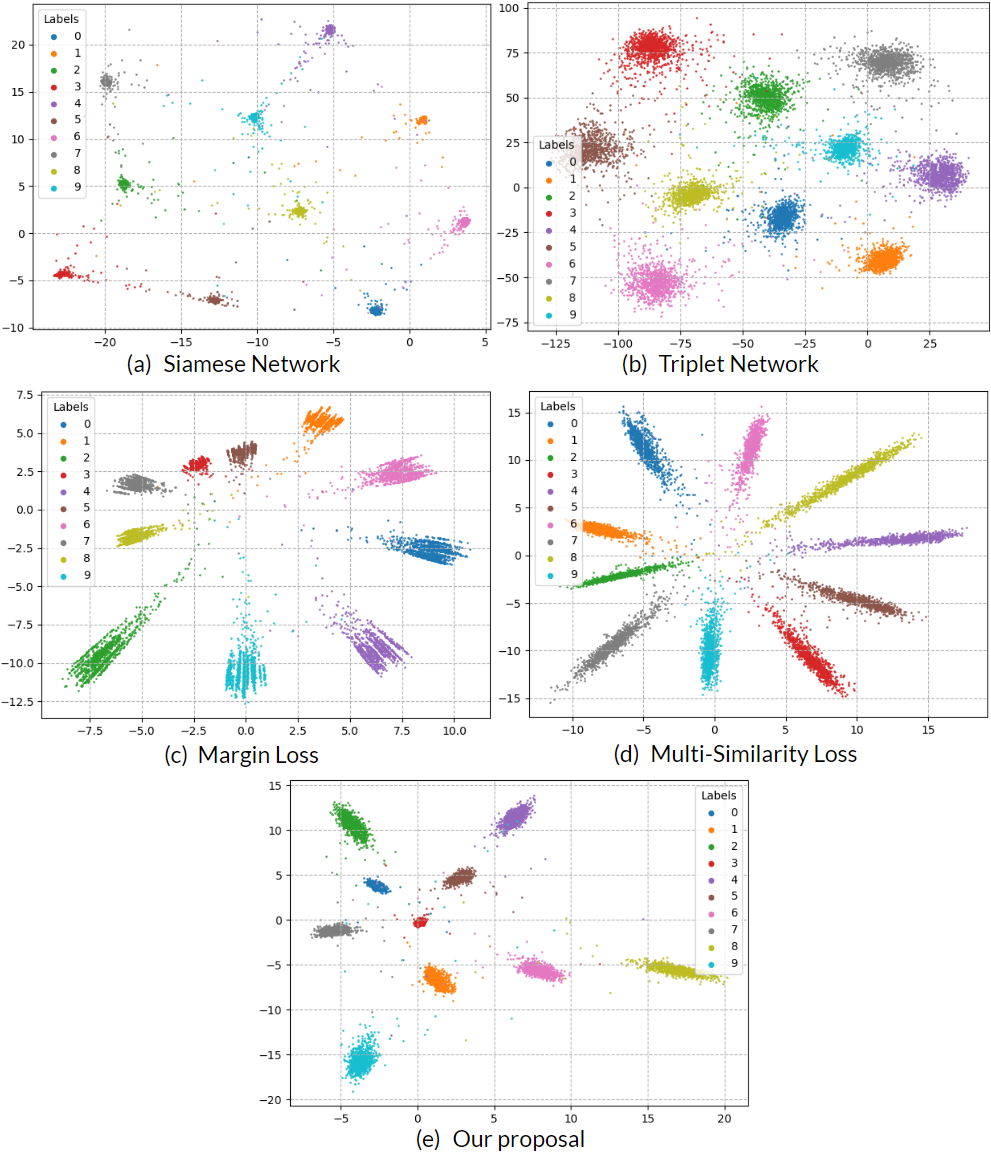}
\caption{Embedding spaces using the MNIST dataset.}
\label{fig:comparison4}
\end{figure}

\begin{figure}[h]
\centering
\captionsetup{justification=justified}
\includegraphics[width=0.496\textwidth]{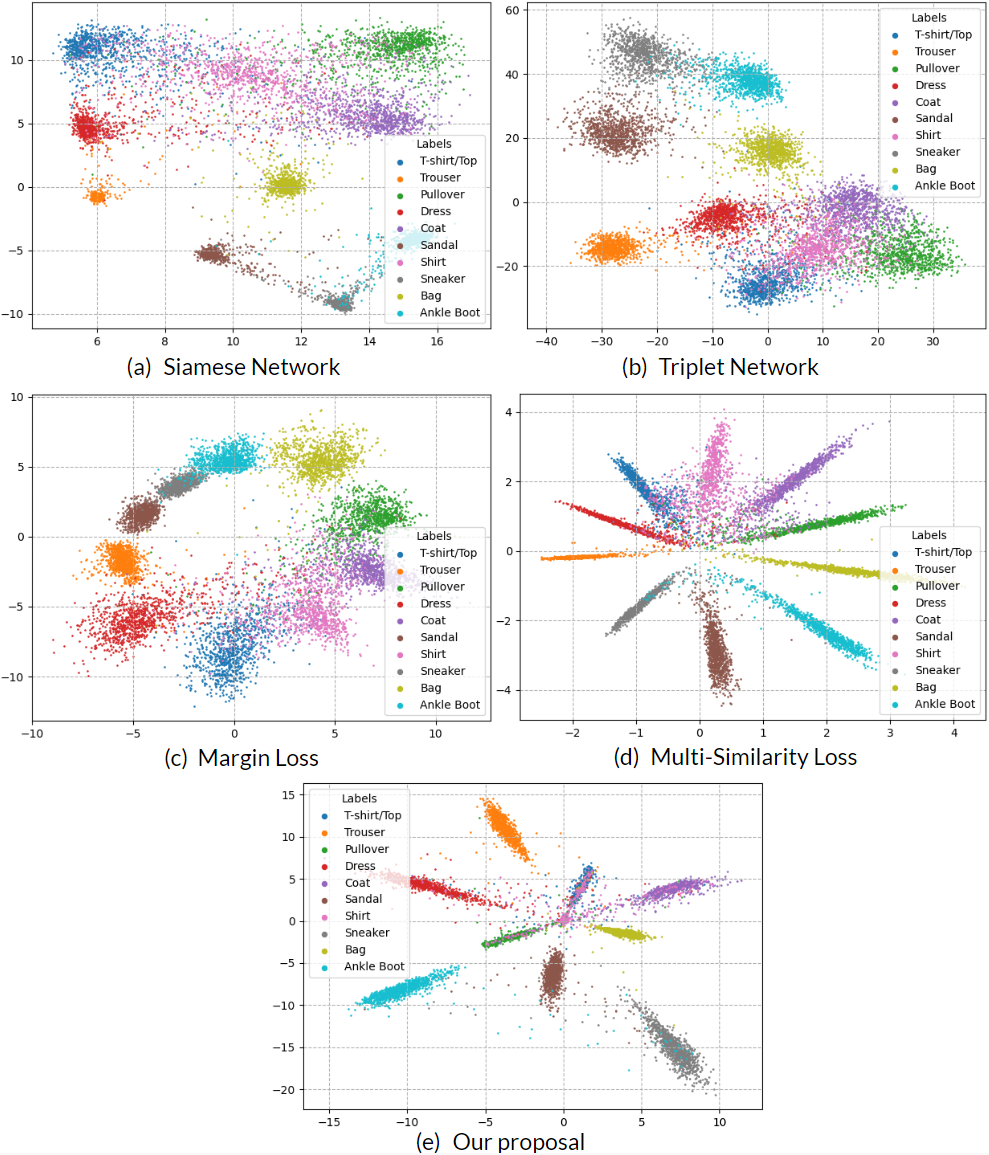}
\caption{Embedding spaces using the Fashion-MNIST dataset.}
\label{fig:comparison5}
\end{figure}

\begin{figure}[h]
\centering
\captionsetup{justification=justified}
\includegraphics[width=0.496\textwidth]{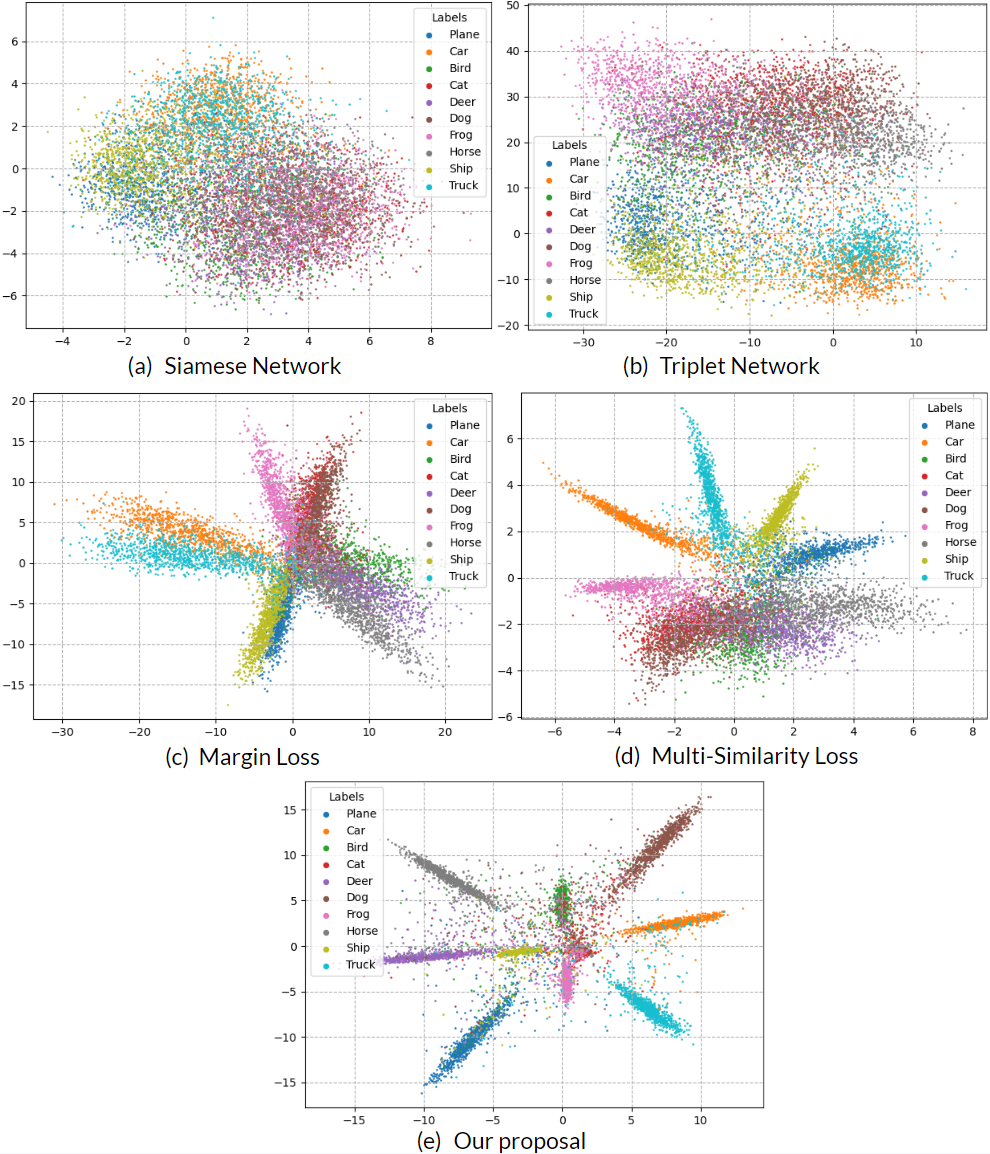}
\caption{Embedding spaces using the CIFAR10 dataset.}
\label{fig:comparison6}
\end{figure}

The choice of the two-dimensional embedding size was motivated to push the proposed method to a more restrictive condition, such as producing low-dimensional outputs, and consequently, to its potential for tasks such as data visualization or more data insight. 

In order to provide more information, figures \ref{fig:comparison4}-\ref{fig:comparison6} show the test set in the embedding spaces learned by the different methods and data sets. We can generally observe that our proposal achieves better compactness and separation of the different classes on the different datasets. It is appreciated particularly in the most difficult dataset (Figure \ref{fig:comparison6}).

\section{Conclusions}
\noindent
In this paper, we analyze Deep Metric Learning (DML) models for embedding learning, the relationships and implementations that connect them to Few-Shot Learning (FSL), and how some models found in FSL have an architecture analogous to knowledge distillation with a change in their approach.

We proposed a DML model that integrates FSL and knowledge distillation concepts to develop an architecture that uses local and global information for better manifold generalization and data representation capabilities, providing better performance and stability compared to the compared methods. It also demonstrates that FSL adaptations boost generalization performance in DML models but also meaningful embeddings can be learned without a strict sample selection phase.

Our approach and the models to which it was compared have been implemented in a careful experimental setup. Seeking the conditions were as similar as possible to avoid unfair comparisons, and we have used metrics that provide a completer picture of the generated embedding space. Though our results were based only on simple datasets, we will perform additional experiments on more closely related datasets with DML or FSL methods and out-of-distribution generalization metrics to further verify the performance.

{\small
\bibliographystyle{ieee_fullname}
\bibliography{egbib}

\begin{thebibliography}{10}\itemsep=-1pt

\bibitem{aghamaleki2019multi}
Javad~Abbasi Aghamaleki and Vahid~Ashkani Chenarlogh.
\newblock Multi-stream cnn for facial expression recognition in limited
  training data.
\newblock {\em Multimedia Tools and Applications}, 78(16):22861--22882, 2019.

\bibitem{benaim2018one}
Sagie Benaim and Lior Wolf.
\newblock One-shot unsupervised cross domain translation.
\newblock {\em advances in neural information processing systems}, 31, 2018.

\bibitem{brattoli2020rethinking}
Biagio Brattoli, Joseph Tighe, Fedor Zhdanov, Pietro Perona, and Krzysztof
  Chalupka.
\newblock Rethinking zero-shot video classification: End-to-end training for
  realistic applications.
\newblock In {\em Proceedings of the IEEE/CVF Conference on Computer Vision and
  Pattern Recognition}, pages 4613--4623, 2020.

\bibitem{chen2018dress}
Long Chen and Yuhang He.
\newblock Dress fashionably: Learn fashion collocation with deep mixed-category
  metric learning.
\newblock In {\em Proceedings of the AAAI Conference on Artificial
  Intelligence}, volume~32, 2018.

\bibitem{chen2018darkrank}
Yuntao Chen, Naiyan Wang, and Zhaoxiang Zhang.
\newblock Darkrank: Accelerating deep metric learning via cross sample
  similarities transfer.
\newblock In {\em Proceedings of the AAAI Conference on Artificial
  Intelligence}, volume~32, 2018.

\bibitem{chenarlogh2019multi}
Vahid~Ashkani Chenarlogh, Farbod Razzazi, and Najmeh Mohammadyahya.
\newblock A multi-view human action recognition system in limited data case
  using multi-stream cnn.
\newblock In {\em 2019 5th Iranian Conference on Signal Processing and
  Intelligent Systems (ICSPIS)}, pages 1--11. IEEE, 2019.

\bibitem{chopra2005learning}
Sumit Chopra, Raia Hadsell, and Yann LeCun.
\newblock Learning a similarity metric discriminatively, with application to
  face verification.
\newblock In {\em 2005 IEEE Computer Society Conference on Computer Vision and
  Pattern Recognition (CVPR'05)}, volume~1, pages 539--546. IEEE, 2005.

\bibitem{deng2019arcface}
Jiankang Deng, Jia Guo, Niannan Xue, and Stefanos Zafeiriou.
\newblock Arcface: Additive angular margin loss for deep face recognition.
\newblock In {\em Proceedings of the IEEE/CVF conference on computer vision and
  pattern recognition}, pages 4690--4699, 2019.

\bibitem{du2022vos}
Xuefeng Du, Zhaoning Wang, Mu Cai, and Yixuan Li.
\newblock Vos: Learning what you don't know by virtual outlier synthesis.
\newblock {\em arXiv preprint arXiv:2202.01197}, 2022.

\bibitem{furlanello2018born}
Tommaso Furlanello, Zachary Lipton, Michael Tschannen, Laurent Itti, and Anima
  Anandkumar.
\newblock Born again neural networks.
\newblock In {\em International Conference on Machine Learning}, pages
  1607--1616. PMLR, 2018.

\bibitem{gou2021knowledge}
Jianping Gou, Baosheng Yu, Stephen~J Maybank, and Dacheng Tao.
\newblock Knowledge distillation: A survey.
\newblock {\em International Journal of Computer Vision}, 129(6):1789--1819,
  2021.

\bibitem{hadsell2006dimensionality}
Raia Hadsell, Sumit Chopra, and Yann LeCun.
\newblock Dimensionality reduction by learning an invariant mapping.
\newblock In {\em 2006 IEEE Computer Society Conference on Computer Vision and
  Pattern Recognition (CVPR'06)}, volume~2, pages 1735--1742. IEEE, 2006.

\bibitem{harwood2017smart}
Ben Harwood, Vijay Kumar~BG, Gustavo Carneiro, Ian Reid, and Tom Drummond.
\newblock Smart mining for deep metric learning.
\newblock In {\em Proceedings of the IEEE International Conference on Computer
  Vision}, pages 2821--2829, 2017.

\bibitem{hinton2015distilling}
Geoffrey Hinton, Oriol Vinyals, and Jeff Dean.
\newblock Distilling the knowledge in a neural network.
\newblock {\em arXiv preprint arXiv:1503.02531}, 2015.

\bibitem{hoffer2015deep}
Elad Hoffer and Nir Ailon.
\newblock Deep metric learning using triplet network.
\newblock In {\em International workshop on similarity-based pattern
  recognition}, pages 84--92. Springer, 2015.

\bibitem{ji2021power}
Wenlong Ji, Zhun Deng, Ryumei Nakada, James Zou, and Linjun Zhang.
\newblock The power of contrast for feature learning: A theoretical analysis.
\newblock {\em arXiv preprint arXiv:2110.02473}, 2021.

\bibitem{kaya2019deep}
Mahmut Kaya and Hasan~{\c{S}}akir Bilge.
\newblock Deep metric learning: A survey.
\newblock {\em Symmetry}, 11(9):1066, 2019.

\bibitem{kim2021embedding}
Sungyeon Kim, Dongwon Kim, Minsu Cho, and Suha Kwak.
\newblock Embedding transfer with label relaxation for improved metric
  learning.
\newblock In {\em Proceedings of the IEEE/CVF Conference on Computer Vision and
  Pattern Recognition}, pages 3967--3976, 2021.

\bibitem{koh2021wilds}
Pang~Wei Koh, Shiori Sagawa, Henrik Marklund, Sang~Michael Xie, Marvin Zhang,
  Akshay Balsubramani, Weihua Hu, Michihiro Yasunaga, Richard~Lanas Phillips,
  Irena Gao, et~al.
\newblock Wilds: A benchmark of in-the-wild distribution shifts.
\newblock In {\em International Conference on Machine Learning}, pages
  5637--5664. PMLR, 2021.

\bibitem{krizhevsky2009learning}
Alex Krizhevsky, Geoffrey Hinton, et~al.
\newblock Learning multiple layers of features from tiny images.
\newblock 2009.

\bibitem{lecun1998gradient}
Yann LeCun, L{\'e}on Bottou, Yoshua Bengio, and Patrick Haffner.
\newblock Gradient-based learning applied to document recognition.
\newblock {\em Proceedings of the IEEE}, 86(11):2278--2324, 1998.

\bibitem{milbich2020diva}
Timo Milbich, Karsten Roth, Homanga Bharadhwaj, Samarth Sinha, Yoshua Bengio,
  Bj{\"o}rn Ommer, and Joseph~Paul Cohen.
\newblock Diva: Diverse visual feature aggregation for deep metric learning.
\newblock In {\em European Conference on Computer Vision}, pages 590--607.
  Springer, 2020.

\bibitem{milbich2021characterizing}
Timo Milbich, Karsten Roth, Samarth Sinha, Ludwig Schmidt, Marzyeh Ghassemi,
  and Bj{\"o}rn Ommer.
\newblock Characterizing generalization under out-of-distribution shifts in
  deep metric learning.
\newblock {\em arXiv preprint arXiv:2107.09562}, 2021.

\bibitem{motiian2017few}
Saeid Motiian, Quinn Jones, Seyed Iranmanesh, and Gianfranco Doretto.
\newblock Few-shot adversarial domain adaptation.
\newblock {\em Advances in neural information processing systems}, 30, 2017.

\bibitem{musgrave2020metric}
Kevin Musgrave, Serge Belongie, and Ser-Nam Lim.
\newblock A metric learning reality check.
\newblock In {\em European Conference on Computer Vision}, pages 681--699.
  Springer, 2020.

\bibitem{musgrave2020pytorch}
Kevin Musgrave, Serge Belongie, and Ser-Nam Lim.
\newblock Pytorch metric learning, 2020.

\bibitem{oh2016deep}
Hyun Oh~Song, Yu Xiang, Stefanie Jegelka, and Silvio Savarese.
\newblock Deep metric learning via lifted structured feature embedding.
\newblock In {\em Proceedings of the IEEE conference on computer vision and
  pattern recognition}, pages 4004--4012, 2016.

\bibitem{rajasegaran2020self}
Jathushan Rajasegaran, Salman Khan, Munawar Hayat, Fahad~Shahbaz Khan, and
  Mubarak Shah.
\newblock Self-supervised knowledge distillation for few-shot learning.
\newblock {\em arXiv preprint arXiv:2006.09785}, 2020.

\bibitem{roth2021simultaneous}
Karsten Roth, Timo Milbich, Bjorn Ommer, Joseph~Paul Cohen, and Marzyeh
  Ghassemi.
\newblock Simultaneous similarity-based self-distillation for deep metric
  learning.
\newblock In {\em International Conference on Machine Learning}, pages
  9095--9106. PMLR, 2021.

\bibitem{roth2020revisiting}
Karsten Roth, Timo Milbich, Samarth Sinha, Prateek Gupta, Bjorn Ommer, and
  Joseph~Paul Cohen.
\newblock Revisiting training strategies and generalization performance in deep
  metric learning.
\newblock In {\em International Conference on Machine Learning}, pages
  8242--8252. PMLR, 2020.

\bibitem{shen2021towards}
Zheyan Shen, Jiashuo Liu, Yue He, Xingxuan Zhang, Renzhe Xu, Han Yu, and Peng
  Cui.
\newblock Towards out-of-distribution generalization: A survey.
\newblock {\em arXiv preprint arXiv:2108.13624}, 2021.

\bibitem{snell2017prototypical}
Jake Snell, Kevin Swersky, and Richard Zemel.
\newblock Prototypical networks for few-shot learning.
\newblock {\em Advances in neural information processing systems}, 30, 2017.

\bibitem{sohn2016improved}
Kihyuk Sohn.
\newblock Improved deep metric learning with multi-class n-pair loss objective.
\newblock In {\em Advances in neural information processing systems}, pages
  1857--1865, 2016.

\bibitem{sung2018learning}
Flood Sung, Yongxin Yang, Li Zhang, Tao Xiang, Philip~HS Torr, and Timothy~M
  Hospedales.
\newblock Learning to compare: Relation network for few-shot learning.
\newblock In {\em Proceedings of the IEEE conference on computer vision and
  pattern recognition}, pages 1199--1208, 2018.

\bibitem{tian2019contrastive}
Yonglong Tian, Dilip Krishnan, and Phillip Isola.
\newblock Contrastive representation distillation.
\newblock {\em arXiv preprint arXiv:1910.10699}, 2019.

\bibitem{tian2020rethinking}
Yonglong Tian, Yue Wang, Dilip Krishnan, Joshua~B Tenenbaum, and Phillip Isola.
\newblock Rethinking few-shot image classification: a good embedding is all you
  need?
\newblock In {\em European Conference on Computer Vision}, pages 266--282.
  Springer, 2020.

\bibitem{wang2017deep}
Jian Wang, Feng Zhou, Shilei Wen, Xiao Liu, and Yuanqing Lin.
\newblock Deep metric learning with angular loss.
\newblock In {\em Proceedings of the IEEE International Conference on Computer
  Vision}, pages 2593--2601, 2017.

\bibitem{wang2019multi}
Xun Wang, Xintong Han, Weilin Huang, Dengke Dong, and Matthew~R Scott.
\newblock Multi-similarity loss with general pair weighting for deep metric
  learning.
\newblock In {\em Proceedings of the IEEE/CVF Conference on Computer Vision and
  Pattern Recognition}, pages 5022--5030, 2019.

\bibitem{wang2020generalizing}
Yaqing Wang, Quanming Yao, James~T Kwok, and Lionel~M Ni.
\newblock Generalizing from a few examples: A survey on few-shot learning.
\newblock {\em ACM Computing Surveys (CSUR)}, 53(3):1--34, 2020.

\bibitem{wu2017sampling}
Chao-Yuan Wu, R Manmatha, Alexander~J Smola, and Philipp Krahenbuhl.
\newblock Sampling matters in deep embedding learning.
\newblock In {\em Proceedings of the IEEE International Conference on Computer
  Vision}, pages 2840--2848, 2017.

\bibitem{xiao2017fashion}
Han Xiao, Kashif Rasul, and Roland Vollgraf.
\newblock Fashion-mnist: a novel image dataset for benchmarking machine
  learning algorithms.
\newblock {\em arXiv preprint arXiv:1708.07747}, 2017.

\bibitem{yang2021generalized}
Jingkang Yang, Kaiyang Zhou, Yixuan Li, and Ziwei Liu.
\newblock Generalized out-of-distribution detection: A survey.
\newblock {\em arXiv preprint arXiv:2110.11334}, 2021.

\bibitem{zhang2019your}
Linfeng Zhang, Jiebo Song, Anni Gao, Jingwei Chen, Chenglong Bao, and Kaisheng
  Ma.
\newblock Be your own teacher: Improve the performance of convolutional neural
  networks via self distillation.
\newblock In {\em Proceedings of the IEEE/CVF International Conference on
  Computer Vision}, pages 3713--3722, 2019.

\bibitem{zhang2018fine}
Yabin Zhang, Hui Tang, and Kui Jia.
\newblock Fine-grained visual categorization using meta-learning optimization
  with sample selection of auxiliary data.
\newblock In {\em Proceedings of the european conference on computer vision
  (ECCV)}, pages 233--248, 2018.

\end{thebibliography}
}

\end{document}